\documentclass{article}

\usepackage{microtype}
\usepackage{hyperref}
\usepackage{url}
\usepackage{booktabs} 
\usepackage{etoolbox,xspace}
\usepackage{hyperref}
\usepackage{amsmath}
\usepackage{amssymb}
\usepackage{mathtools}
\usepackage{amsthm}
\usepackage{wrapfig}
\usepackage{caption}
\usepackage{enumitem}
\usepackage{titlesec}
\usepackage{subfig}
\usepackage{graphicx}
\usepackage{cleveref}
\usepackage{multirow}
\usepackage{colortbl}
\usepackage{subfig} 

\usepackage{soul}

\usepackage{hyperref}
\usepackage{nicematrix} 

\newcommand{\hlc}[2][yellow]{{%
    \colorlet{foo}{#1}%
    \sethlcolor{foo}\hl{#2}}%
}

\usepackage[accepted]{icml2025}

\theoremstyle{plain}

\theoremstyle{definition}

\theoremstyle{remark}

\usepackage[textsize=tiny]{todonotes}

\icmltitlerunning{Optimizing LLM Merging at Scale}

\crefname{figure}{Fig.}{Figs.}
\crefname{section}{Sect.}{Sects.}
\crefname{appendix}{App.}{Apps.}

\begin{document}

\twocolumn[
\icmltitle{ 
\includegraphics[height=0.80em]{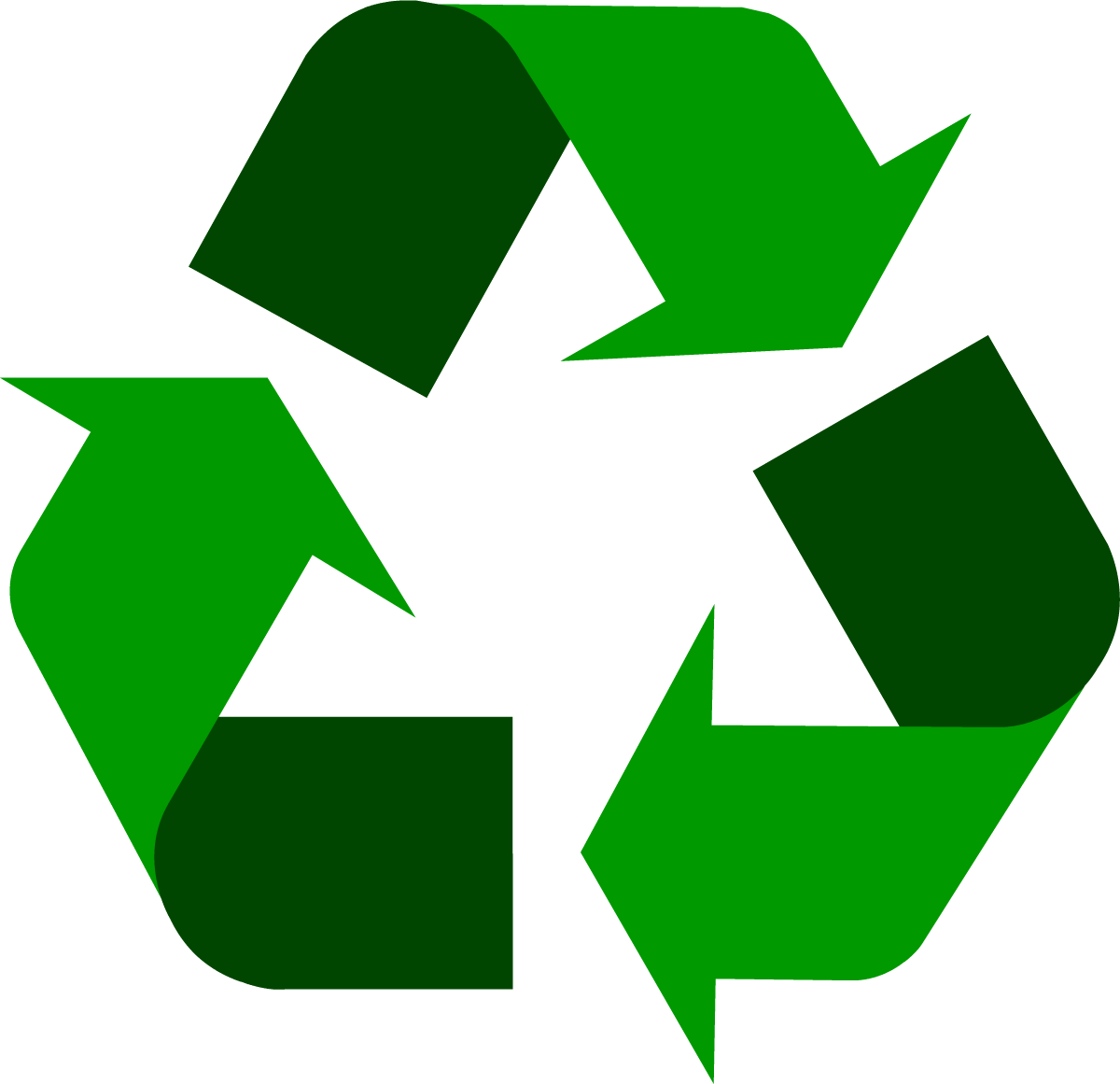} 
%
If You Can't Use Them, Recycle Them: \\ 
Optimizing Merging at Scale Mitigates Performance Tradeoffs 
}



\icmlsetsymbol{equal}{*}

\begin{icmlauthorlist}
\icmlauthor{Muhammad Khalifa}{umich,cohere}
\icmlauthor{Yi-Chern Tan}{cohere}
\icmlauthor{Arash Ahmadian}{c4ai}
\icmlauthor{Tom Hosking}{cohere}
\icmlauthor{Honglak Lee}{umich}
\icmlauthor{Lu Wang}{umich}
\icmlauthor{Ahmet \"Ust\"un}{c4ai}
\icmlauthor{Tom Sherborne}{cohere}
\icmlauthor{Matthias Gall\'e}{cohere}
\end{icmlauthorlist}

\icmlaffiliation{umich}{University of Michigan}
\icmlaffiliation{cohere}{Cohere}
\icmlaffiliation{c4ai}{Cohere For AI}

\icmlcorrespondingauthor{Muhammad Khalifa}{khalifam@umich.edu}

\icmlkeywords{Machine Learning, ICML}

\vskip 0.3in
 ] 



\printAffiliationsAndNotice{}  

\begin{abstract}
Model merging has shown great promise at combining expert models, but the benefit of merging is unclear when merging ``generalist'' models trained on many tasks. We explore merging in the context of large  ($\sim$100B) models, by \textit{recycling} checkpoints that exhibit tradeoffs among different tasks.
Such checkpoints are often created in the process of developing a frontier model, and the suboptimal ones are usually discarded. 
Given a pool of model checkpoints obtained from different training runs (e.g., different stages, objectives, hyperparameters, and data mixtures), which naturally show tradeoffs across different language capabilities (e.g., instruction following vs. code generation), we investigate whether merging can recycle such suboptimal models into a Pareto-optimal one. 
Our optimization algorithm tunes the weight of each checkpoint in a linear combination, resulting in such an optimal model that outperforms both individual models and merge-based baselines. 
Further analysis shows that good merges tend to include almost all checkpoints with non-zero weights, indicating that even seemingly bad initial checkpoints can contribute to good final merges.
\end{abstract}

\section{Introduction}

Model merging is gaining traction as a cost-effective alternative to joint multi-task learning \citep{soups22,yu2024dare}. While the research here has rapidly advanced in the last few years, it remains limited in terms of both \textit{model scale} and \textit{the nature of considered checkpoints}. 
On one hand, most work has studied merging fairly small models ($ \leq 10B$) by today's standards, and it remains unclear how much benefit merging brings with larger models ($ \geq 100B $). On the other hand, the setup where merging was mostly applied involved two or more \textit{expert} models, where experts are independently optimized for specialized tasks, merged to combine their capabilities \citep{ties,yu2024dare,akiba2024evolutionary}. The primary motivation for such expert merging is to eliminate the cost of multi-task training---each expert can be trained separately, and then later merged for combined expertise \citep{li2022branch}. 
However, expert merging is only reasonable when expert models are available.
Departing from that setting, modern large language model (LLM) development scenarios tend to produce a large number of multi-task models.

Training general-purpose LLMs relies on training a single model on many tasks during supervised finetuning (SFT)/instruction tuning \citep{chung2024scaling,mistral,gpt4,llama3,team2024gemma,team2023gemini}. A prevalent issue here is that different tasks may conflict with each other \citep{lin2019pareto}, resulting in \textit{tradeoffs} among different capabilities. Parameters well suited to one task may conflict or combine poorly with parameters specialized for another \citep{gueta-etal-2023-knowledge}.
For example, aligning a model with human preferences can hurt performance on other evaluations \citep{bai2022training,safety-utility}. Similarly, improving math capabilities may come at the cost of other reasoning skills~\citep{specializecot23}. One strategy to alleviate these tradeoffs is to carefully tune different training choices until a good enough (e.g., a Pareto-optimal) model is obtained. 
This approach is not only computationally expensive, but also discards suboptimal models under the assumption that they constitute ``failed" experiments.

\begin{figure*}
    \centering
    \includegraphics[width=.99\linewidth]{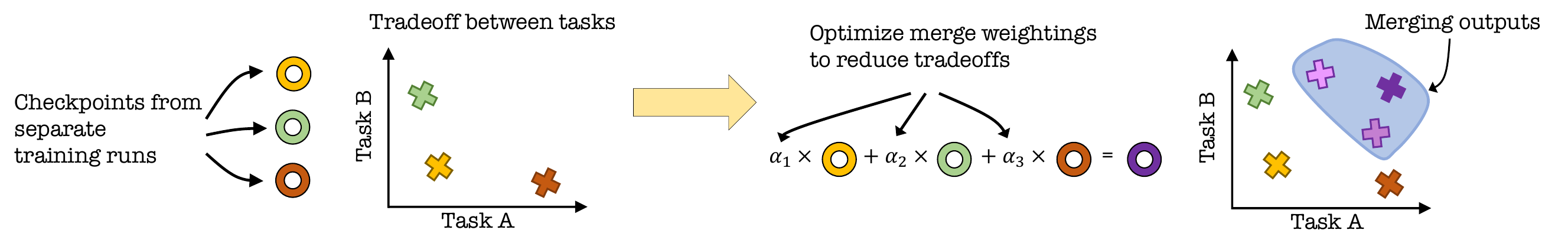}
    \caption{An overview of our setup. Given models obtained from different LLM training runs, we optimize linear merging weightings ($\alpha_1, \alpha_2, \alpha_3$) via iterative search to obtain a model with minimal task tradeoffs. Each \includegraphics[height=0.7em]{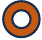} represents a single model, with a \includegraphics[height=0.7em]{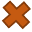} to designate its performance on the two tasks. The purple color indicates a Pareto-optimal model, achieving a good balance between the two tasks without being dominated by other models. We show tradeoffs between only two tasks since it is easier to visualize.}
    \label{fig:overview}
\end{figure*}

Adopting a realistic setting where we use 16 104B checkpoints that exhibit such tradeoffs,
we ask: \textit{How to merge multi-tasked models to reduce performance tradeoffs among different tasks?}
We apply iterative search to optimize the contribution of each checkpoint to the mixture (see Fig.~\ref{fig:overview})

This paper extends model merging approaches by \textbf{(i)} utilizing models orders of magnitudes larger than previous work and, more importantly, \textbf{(ii)} leveraging it to minimize tradeoffs among multiple ``generalist'' checkpoints, as opposed to merging only ``experts''. 
This practical setup considers how to optimally re-use suboptimal intermediate checkpoints in the style of \textit{model recycling}~\citep{choshen2022start}.

We make the following contributions: 
\begin{itemize}[leftmargin=1cm,itemsep=0.1pt]

    \item We explore an understudied setup where the goal is to optimize task tradeoffs by merging multiple models obtained from different training runs, with different data mixtures, post-training stages, and objectives (instead of only random seeds or hyperparamerts). This setting mirrors well the real-life scenario of developing a frontier model (\S\ref{sec:method-task-conflict}). 

    \item We propose to use evolutionary optimization to find such optimal merges instead of relying on manual tuning the weights, a process that can be tedious, unreliable and costly (\S\ref{sec:method-search}). We show that such evolutionary optimization reduces task tradeoffs in two- and three-task setups, outperforming uniform and greedy merging without hurting out-of-domain performance on held-out tasks (\S\ref{sec:results}).
    
    \item We analyze the optimal merges obtained via search and identify key lessons (\S\ref{sec:analysis}): Merges with the best tradeoffs are the result of merging almost all checkpoints, suggesting that linear merging benefits from seemingly \textit{bad} checkpoints that perform poorly on the tasks in question. In other words, merging can be optimized to recycle existing checkpoints, enabling training-free optimization of task tradeoffs. We also observe that the performance of a merge cannot be determined solely based on the isolated performance of the merged models, where seemingly low-performing checkpoints can lead to optimal merges. 

\end{itemize}

\section{Related Work}
Prior work on model merging has mainly considered two axes: merging \textit{how to merge} (methods); and \textit{when to merge} (context).

\paragraph{Merging methods} Recent efforts on model merging methods aim to overcome the limitations of simple averaging \citep{utans1996weight,soups22} with more involved techniques. 
For instance, \citet{matena2022merging} use the Fisher information matrix to approximate each model's posterior beyond the isotropic assumption made by averaging. \citet{tv} propose to merge ``task vectors'' instead of full models. 
\citet{ties} innovated upon averaging by proposing a 3-step process to resolve parameter interference among merged models while \citet{daheim2023model} do so by reducing mismatch in parameter gradients. A major downside of most of these methods is that they require access to gradient information, which becomes expensive as the models scale. 
In addition, other techniques such as DARE \citep{yu2024dare} assume access to a shared base model from which experts are fine-tuned, which is limiting in cases with heterogeneous models as in our setup.

\paragraph{Merging context} Model merging has been mainly exploited in the context of transfer learning~\citep{matena2022merging,jin2022dataless,ties,hammoud2024model,akiba2024evolutionary}, where merging is done across a set of \textit{experts}, specialized models that are fine-tuned separately on different tasks or domains, and which are then merged to combine their capabilities into a single multi-task model \citep{rame2024rewarded}. Merging can also be exploited to ablate on dataset mixtures across domains during pre-training \citep{na2024scalable}.
This does not fully capture typical large-scale LLM training scenarios. In practice, LLMs are simultaneously trained on a multitude of tasks during a single training run---suggesting that training is generally not approached as a multiple single-task optimization problem but rather as a single multi-task optimization problem \citep{ouyang2022training,mistral,chung2024scaling,bai2023qwen,gpt4,team2023gemini}. 
It is more common to have multiple models exhibiting different performance tradeoffs than to have expert models specialized in certain domains. 
\citet{team2024gemma} demonstrate this setup and discuss the tuning of SFT hyperparameters (including training data mixture weightings) to minimize tradeoffs between helpfulness, safety, and faithfulness. Our work further explores optimizing merging across such checkpoints to minimize performance tradeoffs across critical benchmark tasks. 

\paragraph{Contrast with prior work} In a similar spirit to our work, \citet{soups22} average models trained across different runs. However, their approach is applied to computer vision models and, more importantly, their goal is to maximize performance on a single task rather than minimize task tradeoffs. \citet{warp} similarly average LLM policies across different runs to improve alignment reward, but their singular goal is improving reward, contrasting to our exploration of tradeoffs. Their setup is also limited to merging a maximum of 5 smaller models, while we explore merging up to 16 104B checkpoints. \citet{llama3} mention that they average checkpoints trained with different hyperparameters and data mixtures, but they do not provide any additional details. Another adjacent recent work is \citet{akiba2024evolutionary}, using evolutionary optimization techniques to optimize merging across two expert models. They explore parameter-space merging (i.e., merging weights) and data-space merging (i.e., merging information flow across different layers). However, their setup is limited to merging only two 7B models. To summarize, our work sheds light on the underexplored setup where model merging can help recycle existing models beyond the top-few experts. Our work differs from prior work in at least three ways. First, our objective is to minimize performance tradeoffs across different training runs, as opposed to merging expert models. Second, our setup involves merging up to 16 different 104B, making our setup more realistic and aligned with the current scales of LLMs. Third, as opposed to relying on simple averaging \citep{soups22} or tuning the weights manually, we leverage evolutionary search to optimize the merging weightings.

\section{Optimizing LLM Merging}
\subsection{Task Conflict}
\label{sec:method-task-conflict}
Different tasks conflict with each other in required expertise, resulting in performance tradeoffs where improvement observed over some tasks incurs performance degradation on other tasks \citep{lin2019pareto,wang2021understanding}. For instance, in language modeling, aligning a model with human preferences could result in an \textit{alignment tax}, where the model performance degrades on held-out tasks \citep{bai2022training}. 
Another example is that instruction tuning could hurt the performance on tasks such as question answering and reasoning \citep{dou2023art}. It is well known in practice that different decisions related to hyperparameters, training data mixtures, or long-context training yield such tradeoffs. 

One way to minimize these tradeoffs is by carefully tuning the training choices. This involves running several training runs with different hyperparameters and choosing Pareto-optimal runs with minimal tradeoffs \citep{team2024gemma}. However, this is cumbersome and expensive, and becomes less tractable as model parameters increase. We investigate whether it is possible to recycle the models obtained from such ``failed'' runs, which are usually abandoned or discarded altogether. We propose to use training-free model merging as an efficient and compute-minimal strategy to combining models. 

\subsection{The Optimization Problem}
\label{sec:method-optimization}
Given $T$ tasks $\{t_1, \cdots, t_T\}$ and $N$ checkpoints $\{\theta_1, \theta_2, ..., \theta_N\}$ which tradeoff across the tasks, we aim to output a model $\theta_\text{mrg}$ by merging the checkpoints such that all tradeoffs are minimized. We focus on weight interpolation of the model weights, or model soups \citep{soups22}, which yield a merged model by linearly interpolating the model parameters with non-negative weightings. Precisely, the resulting model is computed as a weighted sum of the individual model parameters i.e., $\theta_\text{mrg} = \sum_{i=1}^{N} \alpha_i \theta_i$, where $\alpha_i$ represents the weighting assigned to $\theta_i$, subject to $\sum_{i=1}^{N} \alpha_i = 1$.

We limit our study to linear merging (i.e., model soups) for three reasons. First, linear merging is easy to implement and is generally considered a strong baseline \cite{soups22,ilharco2022patching,ties}. Second, it makes few assumptions about the merged models (e.g., assumes no shared base model \citep{ilharco2022tv}, or access to gradient information \citep{matena2022merging}) enabling easy scaling to large models. Third, 
recent work \citep{yadav2024matters} has shown that different merging methods, including linear merging, may converge approximately to the same performance at large model scales.


Let \( P_t(\theta') \) represent the performance of the model \(\theta'\) on task \( t \). Given a candidate model \(\theta'\), we quantify the performance tradeoffs using a fitness function \( R(\theta') = R(P_{t_1}(\theta'), \cdots, P_{t_T}(\theta'))\) which assesses the balance across tasks to capture all tradeoffs. That is, higher fitness means less severe tradeoffs. Thus, our objective is to find the optimal weightings vector \( a^* = \{\alpha_1^*, \alpha_2^*, \cdots, \alpha_N^*\} \) that minimizes the task tradeoffs and therefore maximizes $R$:
$a^* = \arg \max_{a} \, R(P_{t_1}(\theta_\text{mrg}), \dots, P_{t_T}(\theta_\text{mrg}))$ .

This optimization formulation seeks a balanced performance across tasks by adjusting the weights in $a$.

\paragraph{Fitness Function} We rely on single-objective optimizations, and therefore we require a single score that reflects the performance tradeoff between the given tasks. One choice could be the product of all task metrics or a weighted average of the metrics---if some tasks are considered more important than others.
We will use the \textbf{unweighted macro-average} of task performances of the merging output $\theta_\text{mrg}$, matching the majority of LLM evaluations for comparison to prior work \citep{llama3,team2024gemma}. Over the tasks $t\in \{t_1, \cdots t_T\}$, the fitness function of a model $\theta'$ is computed as $R(\theta') = \frac{1}{T} \sum_{t\in \{t_1, \cdots t_T\}} P_t(\theta')$.\footnote{
A model achieving the best average task performance is necessarily Pareto-optimal, as no other model can dominate it without achieving a better average.}

\subsection{Search Algorithm}
\label{sec:method-search}
There exists a plethora of techniques to solve such problems, including Bayesian Optimization \citep{snoek2012practical}, random search \citep{bergstra2012random}, and genetic algorithms \citep{young2015optimizing,alibrahim2021hyperparameter}. 
For the purpose of our study, we focus on Covariance Matrix Adaptation Evolution Strategy \citep{hansen2001completely}, which has proven its usefulness in hyperparameter optimization \citep[CMA-ES]{loshchilov2016cma}, including in model merging literature \citep{akiba2024evolutionary}, as well as requiring minimal hyperparameter configuration.

CMA-ES is an optimization technique suited for continuous, non-linear optimization. It improves solutions by iteratively sampling candidates from a multivariate normal distribution, with a mean representing the instantaneous optimal solution. At each step, CMA-ES adapts the covariance matrix of the normal distribution over time, based on successful solutions, to assign higher probability to good solutions. CMA-ES is suitable for cases where gradient information is not available or is expensive to obtain. Details of CMA-ES are shown in \Cref{alg:cma-es} in \Cref{app:optimization}.

\section{Experimental Setup}
\label{sec:expermients}

\begin{figure*}
    \centering
    \includegraphics[width=1\linewidth]{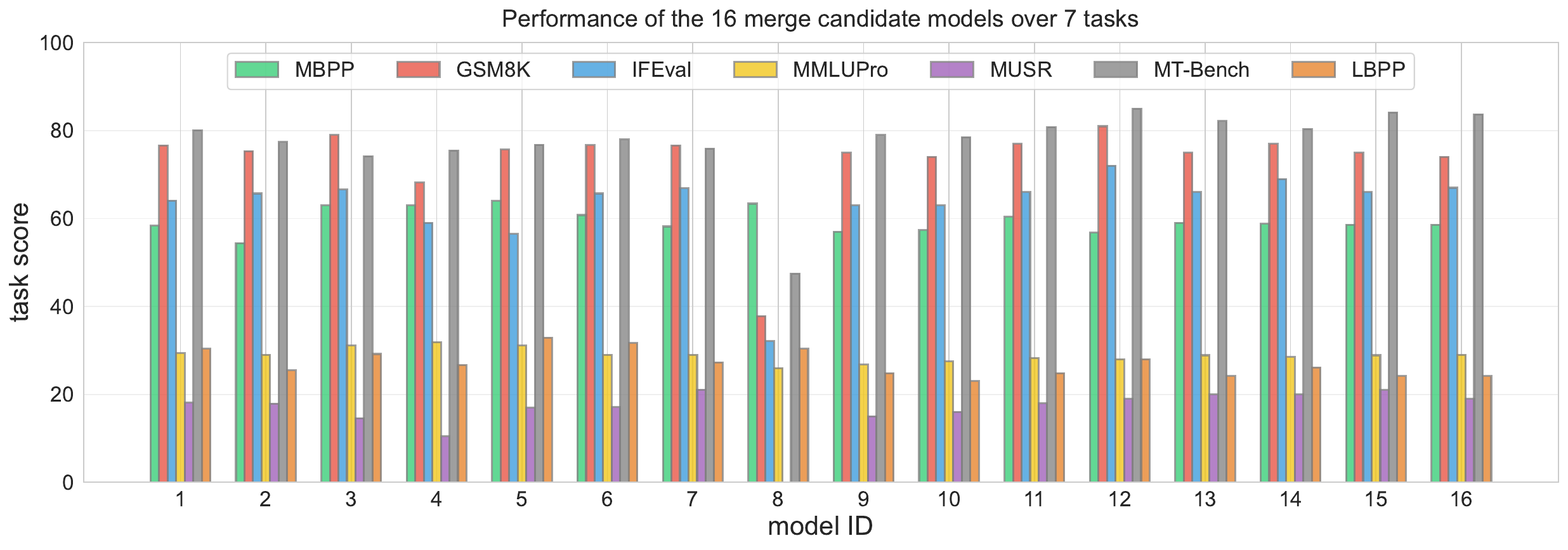}
    \caption{Performance of individual models over the seven tasks covering different capabilities. Models 1-8 are the result of supervised finetuning runs, while 8-16 from preference optimization. Held-out tasks (MT-Bench and LBPP) are used to evaluate the resulting merges to make sure the merge optimization process does not overfit to the held-in tasks that we aim to minimize tradeoffs over. MT-Bench rating is scaled by a factor of 10 for better visualization. The exact numbers are in \Cref{tab:combined-info} in \Cref{app:ckpt-info}.}
    \label{fig:ind-performance}
\end{figure*}

\paragraph{Merge candidates}
As candidate inputs to the merge, we select 16 models from a 104B 
development pipeline, where each model is the result of a separate training run.
To ensure these models are representative of different training stages, we select them such that: 
\begin{itemize}[leftmargin=1cm,itemsep=0.1pt]
    \item 50\% are sourced from the supervised fine-tuning (SFT) training stage, and the other 50\% from preference optimization (PO) stage.
    \item The SFT checkpoints involve different training data: some are trained purely on code, some on code and academic data, and some on the general multi-task SFT setup, etc.,
    \item The PO checkpoints include variation of the training objective and hyperparameters (e.g., warmup ratio).
    \item One checkpoint is sourced from an earlier version of 
    this model's family.
\end{itemize}

\Cref{tab:combined-info} in \Cref{app:ckpt-info} includes details about each model. 
We first note that the PO models are not necessarily based on the SFT models; some are based on different SFT models not included in our candidate models. 
Aside from the criteria above, the checkpoints were selected without any specific bias; we do not intentionally pick checkpoints that exhibit high performance tradeoffs between tasks, as these tradeoffs appear naturally anyway. Our expectation is that our merging strategy can easily apply to other researchers and practitioners who may have many underexploited checkpoints ``laying around''.

\paragraph{Tasks}
We evaluate the merged models on both held-in and held-out tasks. The held-in tasks are the ones for which we wish to minimize the tradeoffs. The held-out tasks serve to verify that the merged model still performs well on tasks outside the held-in tasks. As held-in tasks, we select five different tasks that are representative of different model capabilities. We use \textbf{MBPP}\footnote{\url{https://github.com/google-research/google-research/tree/master/mbpp}.} to evaluate code performance,~\textbf{GSM8K}~\citep{gsm8k} for math,~\textbf{IFEval}~\citep{ifeval} for instruction following,~\textbf{MMLUPro}~\citep{mmlupro}\footnote{To speed up experimentation, we use a subset of 2K examples from MMLU-Pro. We found the performance on this subset to correlate highly with performance on the full dataset.} for multi-task language understanding, and~\textbf{MUSR}~\citep{musr} for multistep reasoning. Held-out tasks include \textbf{MT-bench}~\citep{chatbotarena} and \textbf{LBPP}~\citep{lbpp}.

\paragraph{Evaluation} We report pass@1 for code tasks i.e., MBPP and LBPP. We use accuracy for GSM8K, MMLU Pro, and MUSR. For IFEval, we report the prompt-level strict accuracy \citep{ifeval}, which measures the percentage of prompts for which all verifiable instructions are followed. All evaluations use greedy decoding.




\paragraph{Search details} We use the CMA-ES implementation from Optuna \citep{akiba2019optuna}. 
The population size is set to $4+3\ln N$ (default in the implementation), which aligns with prior work \citep{akiba2024evolutionary}. We initialize the search with uniform weights, i.e., $\alpha_i= 1 / N$ for all $i$. We set the initial standard deviation $\sigma_0 = 1.0$ and run CMA-ES for 50 iterations in total. At each iteration, CMA-ES proposes a single weighting vector $\{\alpha'_1, \alpha'_2, \cdots, \alpha'_N\}$, which we use to merge the individual models and then evaluate the fitness over the resulting merge. To make sure the weightings sum to 1, we first sample unnormalized weights and then normalize by their sum. To avoid cold starting the CMA-ES optimization, we initialize the CMA-ES history with the performance from individual checkpoints results and the two merged baselines detailed below.

\paragraph{Baselines}
To validate how optimal the found solution is, we compare it against the following baselines:
\begin{itemize}[leftmargin=*, itemsep=0.1pt]
    \item \textbf{Best single model:} The individual model with the highest fitness function in addition to the individual model with the highest performance on each task.
    \item \textbf{Uniform soup:} Averaging all checkpoints using equal weights: $\alpha_i = \frac{1}{N}$ \citep{soups22}. 
    \item \textbf{Merge-best:} We average the top-performing checkpoints for each task. Specifically, we merge the checkpoints $\{\theta^*_{t_1}, \cdots, \theta_{t_T}^* \}$ where $\theta_t^* = \arg\max_{\theta} P_t(\theta)$ with uniform weight. This corresponds to assigning the highest weights to the best performing model on each task and zero weights to the remaining models.
\end{itemize}

\section{Results and Discussion}
\label{sec:results}
Before merging any models, we first look at the performance of individual models across different tasks to get a sense of the existing tradeoffs. \Cref{fig:ind-performance} shows the performance of each of the 16 merge candidates over both held-in and held-out tasks. \Cref{tab:combined-info} in \Cref{app:ckpt-info} shows exact numbers. Interesting tradeoffs show up when looking at individual model performances. For instance, SFT models (1-8) exhibit better code performance (i.e., on MBPP and LBPP) compared to PO models (9-16), while PO models seem to perform better than SFT ones on MT-Bench and IFEval. This is a tradeoff likely caused by alignment training, which could hurt some other model capabilities \citep{bai2022training}. 

\begin{figure}
    \centering
    \includegraphics[width=0.68\linewidth]{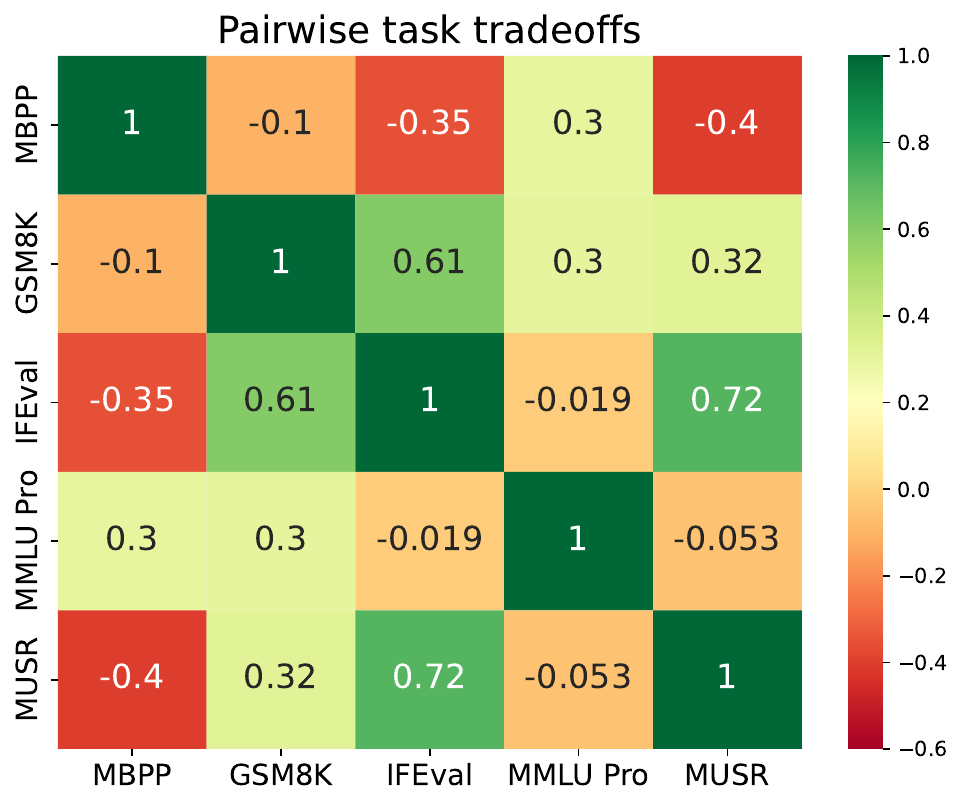}
    \caption{Spearman's rank correlation between task pairs. It is easy to see how some tasks exhibit strong performance tradeoffs, such as MBPP-IFEval and MMLU-Pro/MUSR.}
    \label{fig:corr-pairs}
\end{figure}

Now we zoom in on \textit{pairwise} tradeoffs between two tasks. In this case, it is fairly straightforward to measure the severity of such tradeoffs by computing performance correlation across the models. \Cref{fig:corr-pairs} shows Spearman's rank correlation $\rho$ between all task pairs over our 16 selected models. 
We observe a few strong pairwise tradeoffs between task pairs, such as MBPP-IFEval ($\rho = -0.35$) and MBPP-MUSR ($\rho = -0.40$). 
Throughout this section, we will refer to the tasks with tradeoffs as \textit{held-in} tasks. The main question we aim to answer here through our experiments is: Can optimizing $\{\alpha_1, \alpha_2, \cdots, \alpha_N\}$ yield $\theta_\text{mrg}$ with minimal tradeoffs over the held-in tasks without a hurting performance over the held-out ones?

\begin{figure*}[t!]
    \centering
    \includegraphics[width=0.30\linewidth]{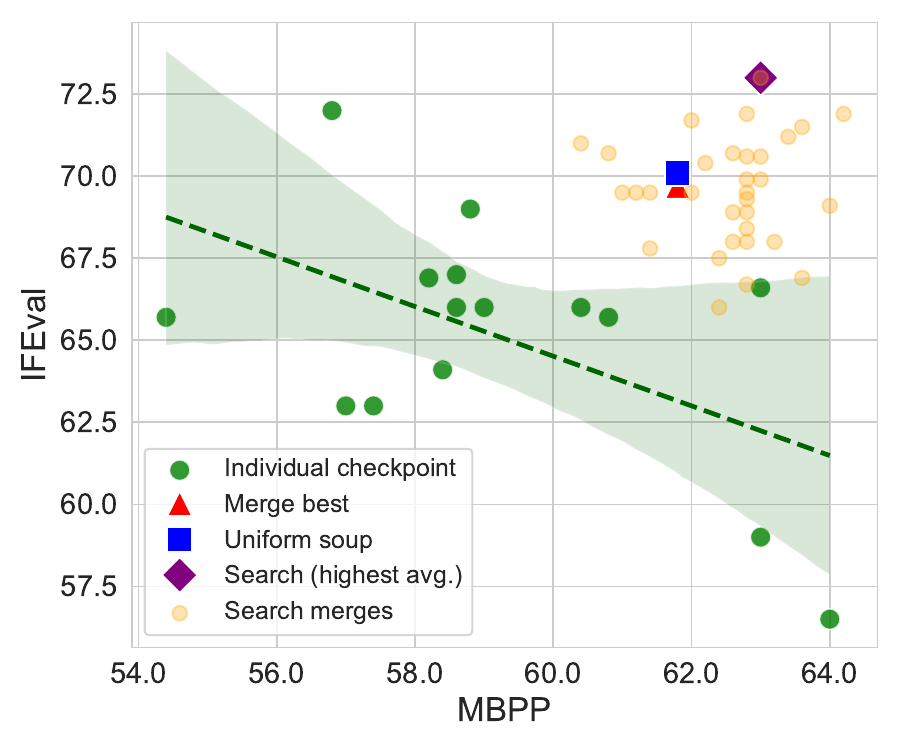}
    \hspace{0.4cm}
    \includegraphics[width=0.30\linewidth]{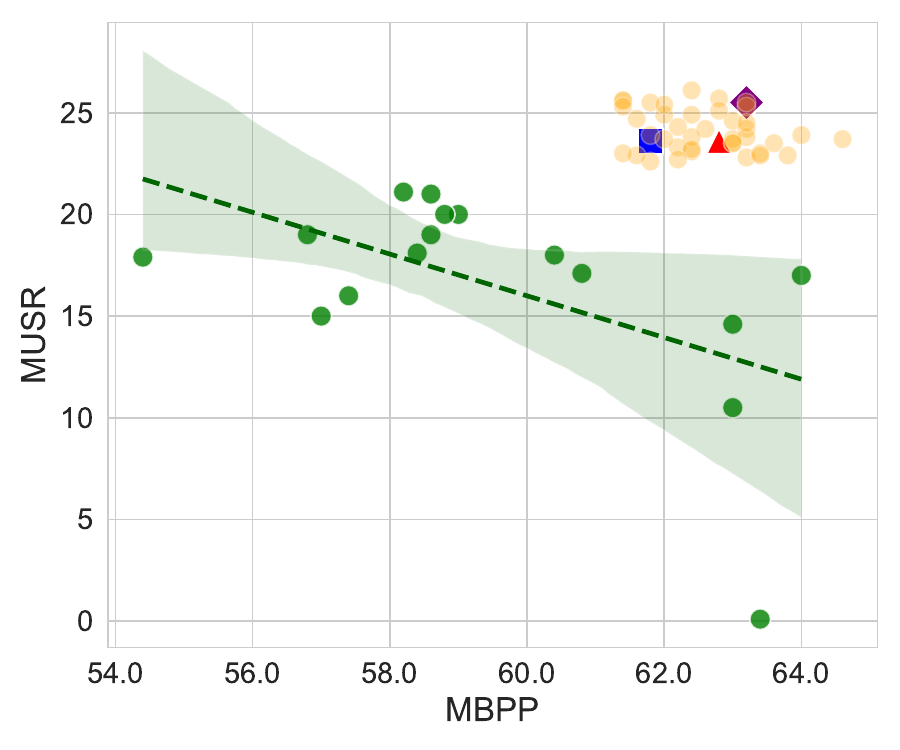}
    \hspace{0.4cm}
    \includegraphics[width=0.30\linewidth]{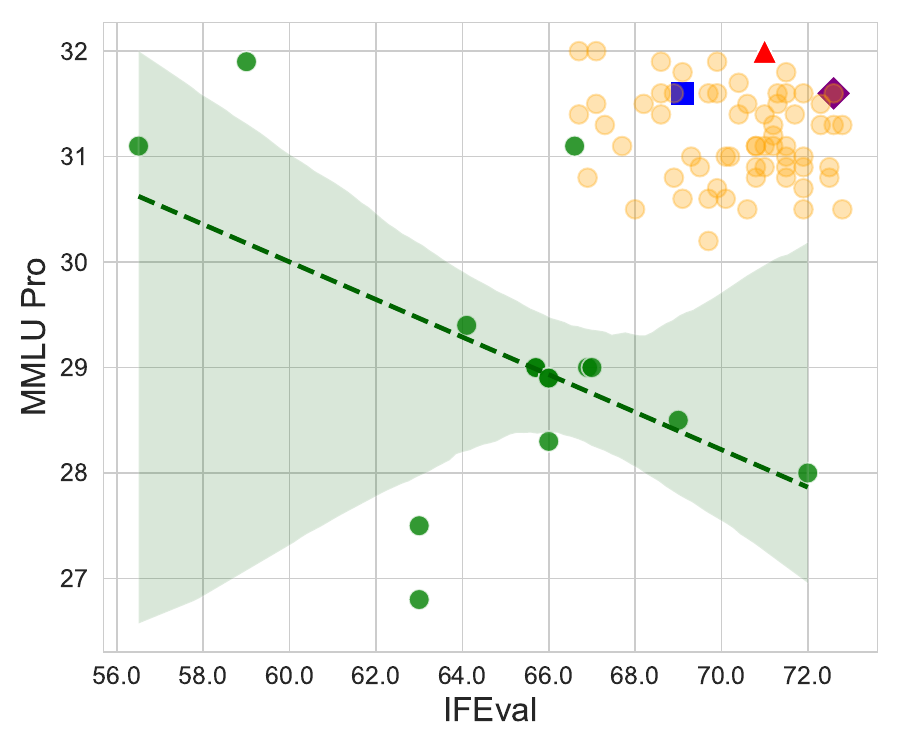}
    \caption{Performance tradeoffs with different merging approaches over different pairwise combinations. Shaded areas represent 95\% confidence interval of the best-fit line computed over individual checkpoint scores (shown in green).}
    \label{fig:pareto-pairs}
\end{figure*}

\begin{table*}[h!]
\centering
\footnotesize
\begin{NiceTabular}{@{}lccccccc@{}}[colortbl-like]
\toprule
\textbf{} & \multicolumn{3}{c}{\textbf{Held-In}}  & \multicolumn{2}{c}{\textbf{Held-Out}} & \textbf{Avg. All Tasks} \\ 
\cmidrule(lr){2-4} \cmidrule(lr){5-6} \cmidrule(lr){7-8}
\textbf{} & \textbf{MBPP} & \textbf{IFEval} & \textbf{Avg.} & \textbf{MT-Bench} & \textbf{LBPP} & \\ 
\midrule
\rowcolor{cyan!15} Highest fitness model & 63.0 & 65.7 & 60.1 & 7.42 & 30.4 &  41.6 \\
\rowcolor{cyan!15} Best on MBPP & 64.0 & 56.5 & 60.3 & 7.68 & 32.9 &  40.3 \\ 
\rowcolor{cyan!15} Best on IFEval & 56.8 & 72.0 & 64.4 & 8.50 & 28.0 &  41.3 \\ 
\midrule
\rowcolor{orange!20} Uniform Soup & 61.8 & 70.1 & 66.0 & 8.13 & 29.8 &  42.5 \\
\rowcolor{orange!20} Merge-best & 61.8 & 69.7 & 65.8 & 8.38 & 32.3 &  43.0 \\ 
\rowcolor{orange!20} Search-optimized (ours) & 63.0 & 73.0 & \textbf{68.0} & 8.07 & 32.3 & \textbf{44.1} \\ 

\midrule
& \textbf{MBPP} & \textbf{MUSR} & \textbf{Avg.} & \textbf{MT-Bench} & \textbf{LBPP}  & &  \\ 
\midrule
\rowcolor{cyan!15} Highest fitness model & 64.0 & 17.0 & 40.5 & 7.68 & 32.9 & 30.4 \\
\rowcolor{cyan!15} Best on MBPP & 64.0 & 17.0 & 40.5 & 7.68 & 32.9 & 30.4 \\
\rowcolor{cyan!15} Best on MUSR & 58.2 & 21.1 & 39.7 & 7.59 & 27.3 & 28.6 \\ 
\midrule
\rowcolor{orange!20} Uniform Soup & 61.8 & 23.6 & 42.7 & 8.17 & 31.7 & 31.3 \\
\rowcolor{orange!20} Merge-best & 62.8 & 23.6 & 43.2 & 7.88 & 30.4 & 31.2 \\ 
\rowcolor{orange!20} Search-optimized (ours)  & 63.2 & 25.5 & \textbf{44.4} & 8.27 & 30.4 & \textbf{31.8} \\

\midrule
& \textbf{MMLU Pro} & \textbf{IFEval} & \textbf{Avg.} & \textbf{MT-Bench} & \textbf{LBPP}  & &  \\  \midrule
\rowcolor{cyan!15} Highest fitness model & 28.0 & 72.0 & 50.0 & 8.50 & 28.0 & 34.1 \\
\rowcolor{cyan!15} Best on MMLUPro & 31.9 & 59.0 & 45.5 & 7.55 & 26.7 & 31.3 \\
\rowcolor{cyan!15} Best on IFEval & 28.0 & 72.0 & 50.0 & 7.55 & 26.7 & 33.6 \\
\midrule
\rowcolor{orange!20} Uniform Soup & 31.6 & 70.1 & 50.9 & 8.19 & 29.2 & 34.8 \\
\rowcolor{orange!20} Merge-best & 32.0 & 71.0 & 51.5 & 8.13 & 31.1 & 35.6 \\
\rowcolor{orange!20} Search-optimized (ours) & 31.6 & 72.6 & \textbf{52.1} & 8.15 & 31.1 & \textbf{35.9}\\

\bottomrule
\end{NiceTabular}
\caption{Performance of different baselines compared to search optimized merge in the pairwise case. Held-in tasks refer to tasks included in the fitness function (\S~\ref{sec:method-optimization}) and held-out tasks are used to validate the quality of the search optimized models. Search yields models with the lowest tradeoffs over the held-in tasks without sacrificing performance on held-out tasks. We highlight \hlc[cyan!25]{single models} and \hlc[orange!20]{merges} differently. Evaluations on MMLU Pro use only 2K test examples.}
\label{tab:pairwise-merge-results}
\end{table*}


\subsection{Optimizing Pairwise Tradeoffs}
\label{sec:pairwise-tradeoffs}

We apply our merge optimization recipe over three task pairs with relatively strong tradeoffs: \textbf{MBPP-IFEval}, \textbf{MBPP-MUSR}, and \textbf{MMLU Pro-IFEval}.

\Cref{fig:pareto-pairs} shows a plot for each pair of tasks. Each plot includes a best-fit line to the performances of each pair (shown in green), which exhibits a negative slope in all three cases due to the respective tradeoff. We observe that baselines such as `Uniform Soup' and `Merge-Best' seem to reduce tradeoffs in a few cases. However, search-based optimization of the merge always yields a model that falls on the Pareto frontier---occasionally outperforming baselines by up to 2.2 average points, as with MBPP-IFEval. In addition, we can see that over MBPP-IFEval and MBPP-MUSR, our search has yielded a model that is better than the best individual model on both IFEval (+1.0) and MUSR (+4.4), showing that search optimized merges could improve over the initial candidate models. We also note that `Merge-Best' baseline, which averages the two best performing models on each task, performs comparably well in some cases (MMLU Pro-IFEval) but performs below search in the other task pairs. 
This suggests that merging based on individual model performance sometimes results in suboptimal merges. 
We dive deeper into the weightings found via search in \Cref{sec:analysis}. 


Looking at the held-in tasks alone however does not give us the full picture. 
It is possible that search has overfit to the held-in tasks by yielding merges with minimal tradeoffs, but could perform significantly worse on tasks that were not incorporated into the fitness function, i.e., out-of-distribution tasks. To verify whether this is the case, we evaluate the resulting merge on held-out tasks, namely, MT-Bench and LBPP. As shown in \Cref{tab:pairwise-merge-results} the search-optimized merges exhibit comparable performance on the held-out tasks---and in some cases even better than baselines. This means search has minimized task tradeoffs over the held-in tasks without compromising performance on other tasks.

\subsection{Optimizing Three-task Tradeoffs}
\label{sec:three-task}
In practice, production LLMs are expected to be performant at more than two tasks. 
We consider balancing performance across three tasks: code generation, instruction following, and math reasoning, by using \textbf{MBPP-IFEval-GSM8K} as held-in tasks. 
We target this combination since IFEval correlates both negatively with MBPP, and positively with GSM8K (as shown in \Cref{fig:corr-pairs}), making it a challenging combination to optimise. 
The goal is to identify how our approach scales with more than 2 tasks, due to the exponential growth in choices (and search space) with respect to the number of tasks.

\begin{figure}[t!]
    \centering
    \includegraphics[width=0.90\linewidth]{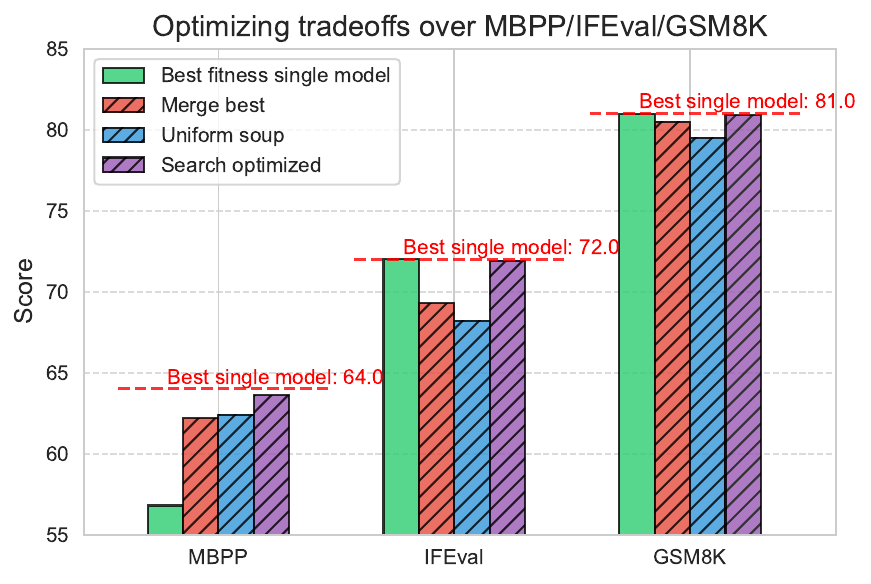}
    \caption{Performance of different merge approaches when minimizing the tradeoffs across three tasks: MBPP, IFEVal, and GSM8K. \textcolor{red}{Dashed red} lines represent the best individual model at the corresponding task. It is clear that search-optimized merging can well balance the performance over the three tasks. Bars corresponding to merging are hatched to differentiate from individual models.}
    \label{fig:three-task-bar}
\end{figure}

Looking at \Cref{fig:three-task-bar}, it is evident that the best fitness single model (i.e., highest average performance) performs well on IFEval and GSM8K, but comparably poor on MBPP. The other two baselines, `Merge-Best' and `Uniform Soup', were able to improve the tradeoffs by some degree but exhibit noticeable performance drop on IFEval. 
While the search-optimized merge is a Pareto-optimal model -- maintaining -- and only slightly underperforms the best single model for each task. These results suggest that search-optimized merging can perform well when scaling the number of tasks. Our approach maintains comparable performance on the held-out tasks, and even improves performance on LBPP compared to the baselines as shown in \Cref{tab:triwise} in \Cref{app:add-results}.


\subsection{Analysis}
\label{sec:analysis}
\begin{figure*}[ht!]
    \centering
    \includegraphics[width=.32\linewidth]{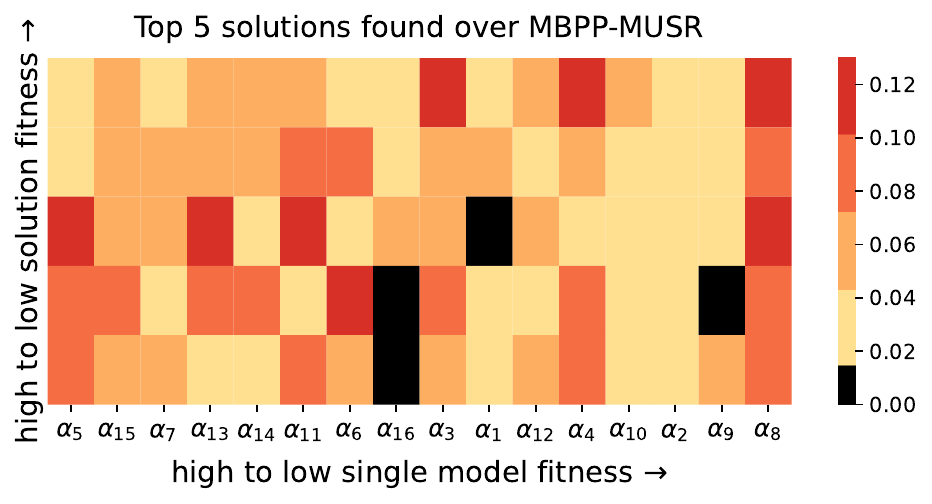}
    \includegraphics[width=.32\linewidth]{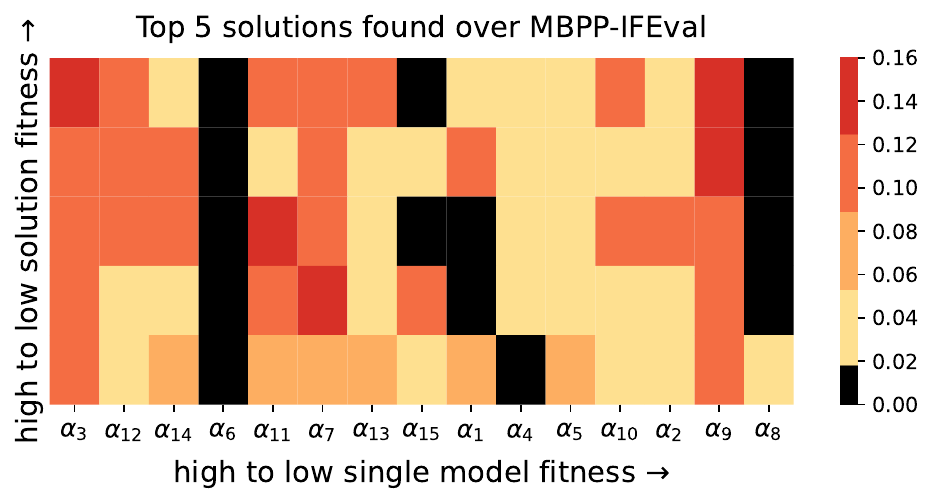}
    \includegraphics[width=.32\linewidth]{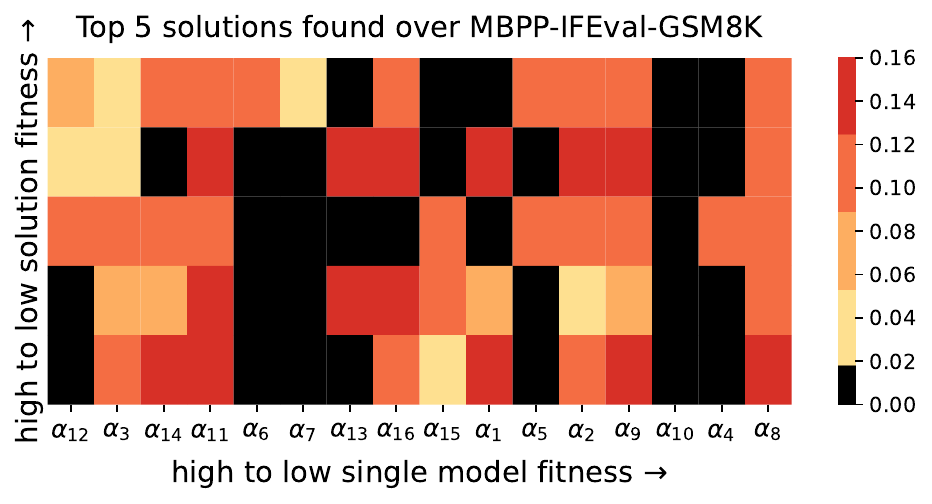}
    \caption{Best solutions found via CMA-ES search when optimizing tradeoffs over the pairs MBPP-MUSR (left) MBPP-IFEval (mid) and MBPP-IFEval-GSM8K. We order the weightings over the x-axis based on the fitness of the individual model they correspond to. We observe that top-solutions do not necessarily assign high weights to high-fitness individual checkpoints. For instance, the top solution on MBPP-IFEval assigns considerably high weight to model \#9, which exhibits a relatively bad tradeoff on the task pair.}
    \label{fig:top-soltuions}
\end{figure*}

We perform further analysis into the dynamics of linear merging:
\paragraph{Most models contribute to the best merges.}
The first question we ask is: What fraction of the initial 16 models contribute to the best solutions? 
\Cref{fig:top-soltuions} shows a heatmap of the top 5 solutions on each task combination. 
We observe that CMA-ES identifies good solutions which distribute the weightings among almost all checkpoints (dense solution), instead of assigning high weights to a small subset of the models (sparse). 
For example, the top solution assigns very few zero weightings (shown in black in \Cref{fig:top-soltuions}) for MBPP-MUSR and MBPP-IFEval. 
Also, while the top solutions for MBPP-IFEVal-GSM8K are slightly sparser in the pairwise case, at least 9/16 weightings are non-zero for the top 5 solutions. 
This indicates that \textbf{almost all checkpoints have contributed to the optimal merge}.

\paragraph{Low performing models may lead to optimal merges.}

Since the CMA-ES optimization process evaluates many solutions, we can investigate the solutions found through search along with their quality, as measured by their stand-alone performance. 
Intuitively, one would expect that high fitness solutions found through search will assign higher weights to the checkpoints that perform well on the held-in tasks, compared to checkpoints that perform poorly. 

Interestingly, this \textit{is not} necessarily the case for the top solutions found by CMA-ES. 
For instance, the top performing solution for MBPP-MUSR has a relatively high weight for $\alpha_8 = 0.09$, even if model \#8 performs extremely poorly at MUSR.\footnote{In fact, this particular code trained model has 0\% accuracy, as shown in \Cref{fig:ind-performance}; It responds to all queries by generating code.}
The same holds for MBPP-IFEval pair, where the top solution does not assign a particularly high weight to any of the best performing models on MBPP (models \#3, \#4, or \#5). 
Similarly, over MBPP-IFEval-GSM8K, $\alpha_8$ is relatively high in the top solution, while model \#8 actually performs the worst on GSM8K. 
\textbf{An individual model's performance on given task(s) does not reflect the performance of a merge that assigns high/low weight to this model}. 
An optimization procedure to find good merges is therefore necessary, since simply assigning the weightings based on the model's isolated performance is suboptimal. 

Similarly, an individual model performance across all held-in tasks is not predictive of its importance in the found solution.
In \Cref{fig:top-soltuions}, we sort the $\alpha_i$ weightings based on the fitness of the corresponding model for each experiment. 
For example, in the leftmost heatmap corresponding to MBPP-MUSR, model \#5 has the highest average performance and model \#8 has the lowest.
If the performance of an individual model would be indicative of its weight in the final merge, we would expect higher weights (more reddish) to be concentrated on the left of each heatmap.
The fact that this does not happen suggests that hand tuning the merge weightings based on heuristics (e.g., individual model performance, fitness, etc.) is suboptimal, further supporting the perspective of approaching model merging as an optimization problem.

\begin{figure}[t!]
    \centering
    \includegraphics[width=0.85\linewidth]{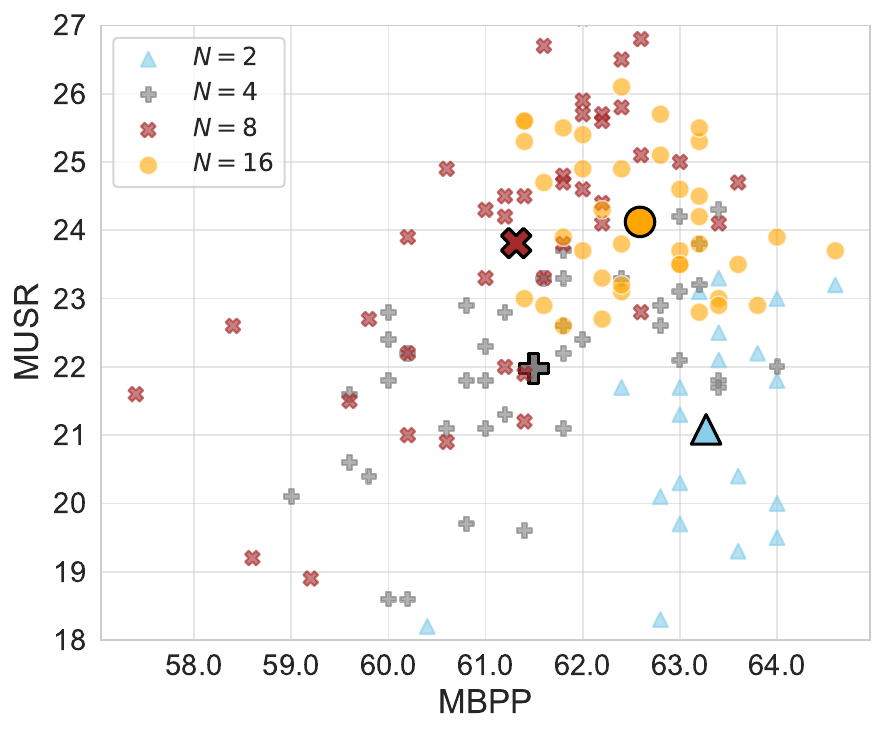}
    \caption{Merges found via CMA-ES when optimizing MBPP-MUSR tradeoffs over 2, 4, 8, and 16 checkpoints. We also show the centroid of each set of experiments (in large markers). Optimizing over more checkpoints (8 and 16) tends to yield less tradeoffs compared to fewer checkpoints (2,4), showing how recycling more models can outperform recycling fewer checkpoints.}
    \label{fig:n_ckpts_exp_mbpp_musr}
\end{figure}

\paragraph{Fitness improves with more iterations.}
We inspect whether CMA-ES effectively optimizes the fitness function as the search progresses by looking at the fitness function development over the course of search. \Cref{fig:search-prog} in \Cref{app:add-results} in  plots the fitness function vs the number of iterations.
Each point in the graph is a weightings vector proposed by CMA-ES, and the fitness is the average of the held-in task performances of the resulting merge. Over the three pairwise task combinations, it is clear that the average solution fitness improves with more CMA-ES iterations. 

\paragraph{Recycling benefits from more checkpoints.}
Including more initial checkpoints obviously extends the search space, but may lead to better solutions.
To study how the merge quality changes with the number of checkpoints, we run CMA-ES over the $N$ checkpoints with the best fitness scores, with $N\in\{2,4,8,16\}$. 
The results over MBPP-IFEval, and MBPP-MUSR are shown in \Cref{fig:n_ckpts_exp_mbpp_musr} and \Cref{fig:n_ckpts_exp_mbpp_ifeval} (in \Cref{app:add-results}), where we highlight the centroid for each $N$.
As can be seen, with larger $N$ the search space is explored more exhaustively and results in centroid checkpoints with better fitness.

\paragraph{Computational cost.}
In \Cref{app:search-cost}, we estimate the computational cost required during both SFT and PO stages of a single model and compare it to the cost of our merge optimization recipe. We find that our recipe requires only about 10\% compute of that needed for training, showing that search-optimized merging provides a significantly cheaper and training-free approach for reducing task tradeoffs.



\section*{Conclusion}
In this paper, we present an approach to recycle checkpoints obtained during a typical training run of a frontier model.
While the vast majority of those checkpoints are in general discarded, in this paper we show how to leverage them via search-optimized merging. 
We show that a simple search algorithm focusing on linear merging can yield better, and often Pareto-optimal models with respect to the existing checkpoints. 
Our research show that it is possible to leverage merging when we have many multi-task trained checkpoints, as opposed to the standard setup of merging experts. 
A surprising finding is that even checkpoints which perform relatively bad on subtasks can contribute to an overall better model.
While we relied on a simple merging approach, we hope future development will further investigate more involved merging techniques in a similar setup to ours via merging as a cheaper and training-free approach.



\section*{Impact Statement}
This paper presents work whose goal is to utilize suboptimal large language model checkpoints by merging them into better models. Potential societal
consequences of our work include accelerating frontier model training and development. There may be other potential societal
consequences of our work, none which we feel must be specifically highlighted here.

\bibliography{ref}
\bibliographystyle{icml2025}

\newpage
\appendix

\newcommand{\Comment}[1]{\hfill\textit{// #1}}
\newcommand{\Return}[1]{\textbf{return} #1}

\section{Optimization Algorithm}
\label{app:optimization}
\Cref{alg:cma-es} Shows a pseudocode of CMA-ES used to optimize the merge weightings, as explained in \Cref{sec:method-optimization}.

\begin{algorithm}[H]
\footnotesize
  \caption{Covariance Matrix Adaptation Evolution Strategy (CMA-ES)}
  \label{alg:cma-es}
  \begin{algorithmic}[1]
    \STATE \textbf{Input:} Objective function $f$, population size $\lambda$, initial mean $\mathbf{m}_0$, initial step size $\sigma_0$, maximum iterations $T$ \label{op1}
    \STATE \textbf{Output:} Optimized solution $\mathbf{m}_\text{opt}$ \label{op2}
    \STATE \textbf{Initialize:} \label{op3}
    \STATE $\mathbf{m} \gets \mathbf{m}_0$ \Comment{Initial mean} \label{op4}
    \STATE $\sigma \gets \sigma_0$ \Comment{Initial step size} \label{op5}
    \STATE $\mathbf{C} \gets \mathbf{I}$ \Comment{Initial covariance matrix} \label{op6}
    \STATE $\mathbf{p}_\sigma \gets \mathbf{0}$, $\mathbf{p}_c \gets \mathbf{0}$ \Comment{Evolution paths} \label{op7}
    \STATE Define recombination weights $w_i$ for $\lambda$ offspring \label{op8}
    \FOR{$t = 1$ to $T$} \label{op9}
        \STATE \textbf{Sample offspring:} \label{op10}
        \FOR{$k = 1$ to $\lambda$} \label{op11}
            \STATE $\mathbf{x}_k \sim \mathcal{N}(\mathbf{m}, \sigma^2 \mathbf{C})$ \label{op12}
        \ENDFOR
        \STATE \textbf{Evaluate fitness:} \label{op13}
        \FOR{$k = 1$ to $\lambda$} \label{op14}
            \STATE $f_k \gets f(\mathbf{x}_k)$ \label{op15}
        \ENDFOR
        \STATE \textbf{Sort offspring by fitness:} \label{op16}
        \STATE Sort $\{\mathbf{x}_k\}$ by ascending $f_k$ \label{op17}
        \STATE \textbf{Update mean:} \label{op18}
        \STATE $\mathbf{m} \gets \sum_{i=1}^\mu w_i \mathbf{x}_{i:\lambda}$ \Comment{$\mu$ best offspring} \label{op19}
        \STATE \textbf{Update evolution paths:} \label{op20}
        \STATE $\mathbf{p}_\sigma \gets (1 - c_\sigma) \mathbf{p}_\sigma + \sqrt{c_\sigma (2 - c_\sigma) \mu_\text{eff}} \mathbf{C}^{-1/2} \frac{\mathbf{m} - \mathbf{m}_\text{prev}}{\sigma}$ \label{op21}
        \STATE $\mathbf{p}_c \gets (1 - c_c) \mathbf{p}_c + \sqrt{c_c (2 - c_c) \mu_\text{eff}} \frac{\mathbf{m} - \mathbf{m}_\text{prev}}{\sigma}$ \label{op22}
        \STATE \textbf{Update covariance matrix:} \label{op23}
        \STATE $\mathbf{C} \gets (1 - c_1 - c_\mu) \mathbf{C} + c_1 \mathbf{p}_c \mathbf{p}_c^\top + c_\mu \sum_{i=1}^\mu w_i (\mathbf{x}_{i:\lambda} - \mathbf{m})(\mathbf{x}_{i:\lambda} - \mathbf{m})^\top$ \label{op24}
        \STATE \textbf{Update step size:} \label{op25}
        \STATE $\sigma \gets \sigma \cdot \exp\left(\frac{c_\sigma}{d_\sigma} \left(\frac{\|\mathbf{p}_\sigma\|}{\mathbb{E}[\|\mathcal{N}(0, \mathbf{I})\|]} - 1 \right)\right)$ \label{op26}
    \ENDFOR
    \STATE \Return $\mathbf{m}_\text{opt} \gets \mathbf{m}$ \label{op27}
  \end{algorithmic}
\end{algorithm}

\section{Checkpoint details}
\label{app:ckpt-info}
We show the details and task performance of the 16 individual checkpoint we use in the paper in \cref{tab:combined-info}.
\begin{table*}[ht]
\footnotesize
\centering
\begin{tabular}{@{}c|p{3cm}|ccccccc@{}}
\toprule
\textbf{Model ID} & \textbf{Info} & \textbf{MBPP} & \textbf{GSM8K} & \textbf{IFEval} & \textbf{MMLUPro} & \textbf{MUSR} & \textbf{MT-Bench} & \textbf{LBPP} \\ \midrule
\multicolumn{9}{@{}l}{\textbf{\textit{Supervised Finetuning}}} \\
1   & Without MuP                                  & 58.4 & 76.6 & 64.1 & 29.4 & 18.1 & 8.01 & 30.4 \\ 
2   & Two stage SFT                                & 54.4 & 75.3 & 65.7 & 29.0 & 17.9 & 7.74 & 25.5 \\ 
3   & Academic + Code data only. 2 epochs         & 63.0 & 79.0 & 66.6 & 31.1 & 14.6 & 7.42 & 29.2 \\ 
4   & Academic + Code data only.                  & 63.0 & 68.2 & 59.0 & 31.9 & 10.5 & 7.55 & 26.7 \\ 
5   & Academic + Code data only. 2 epochs         & 64.0 & 75.7 & 56.5 & 31.1 & 17.0 & 7.68 & 32.9 \\ 
6   & Two stage SFT                                & 60.8 & 76.7 & 65.7 & 29.0 & 17.1 & 7.80 & 31.7 \\ 
7   & Two stage SFT                                & 58.2 & 76.6 & 66.9 & 29.0 & 21.1 & 7.59 & 27.3 \\ 
8   & Only Code                                    & 63.4 & 37.8 & 32.1 & 26.0 & 0.0  & 4.75 & 30.4 \\ \midrule
\multicolumn{9}{@{}l}{\textbf{\textit{Preference Optimization}}} \\ 
9   & Light offline Pref                          & 57.0 & 75.0 & 63.0 & 26.8 & 15.0 & 7.90 & 24.8 \\ 
10  & Data-filtered. Offline Pref                  & 57.4 & 74.0 & 63.0 & 27.5 & 16.0 & 7.85 & 23.0 \\ 
11  & Different Preamble                          & 60.4 & 77.0 & 66.0 & 28.3 & 18.0 & 8.08 & 24.8 \\ 
12  & Light offline Pref, with different margin scaling & 56.8 & 81.0 & 72.0 & 28.0 & 19.0 & 8.50 & 28.0 \\ 
13  & Full offline Pref                           & 59.0 & 75.0 & 66.0 & 28.9 & 20.0 & 8.22 & 24.2 \\ 
14  & Specific data mix. Offline Pref             & 58.8 & 77.0 & 69.0 & 28.5 & 20.0 & 8.04 & 26.1 \\ 
15  & Specific data mix. Offline Pref. With warmup & 58.6 & 75.0 & 66.0 & 28.9 & 21.0 & 8.41 & 24.2 \\ 
16  & Specific data mix. Offline Pref             & 58.6 & 74.0 & 67.0 & 29.0 & 19.0 & 8.37 & 24.2 \\ 
\bottomrule
\end{tabular}
\caption{Details and performance of the different initial checkpoints used for merging in our experiments. Supervised Finetuning and Preference Optimization models are shown with their respective performance across various benchmarks.}
\label{tab:combined-info}
\end{table*}

\section{Additional Results and Analysis}
\label{app:add-results}

\begin{figure}[h!]
    \centering
    \includegraphics[width=0.85\linewidth]{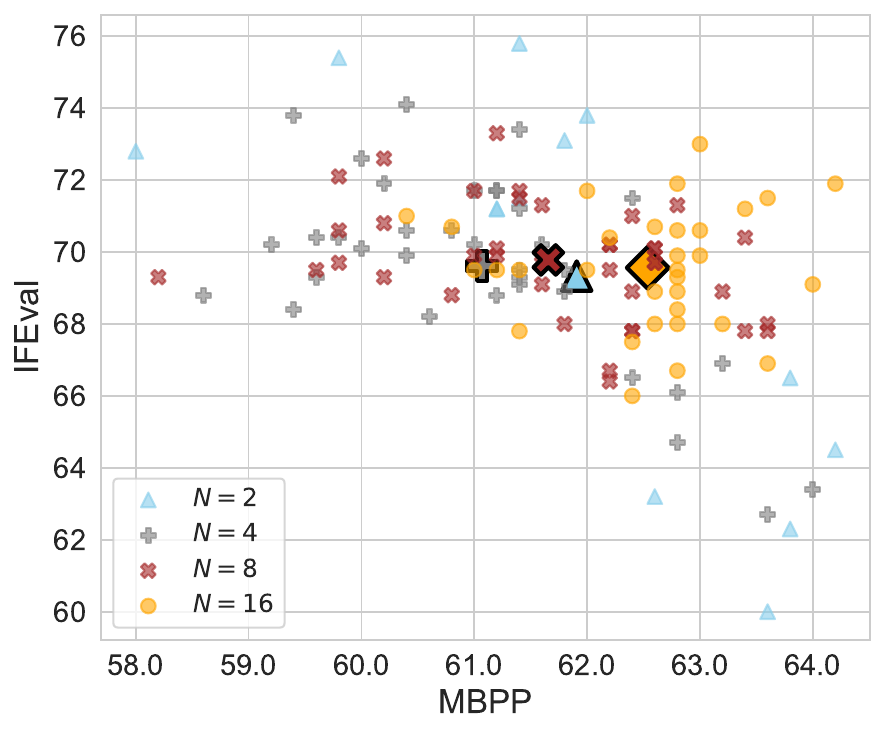}
    \caption{Merges found via CMA-ES when optimizing MBPP-MUSR tradeoffs over 2, 4, 8, and 16 checkpoints. We also show the centroid of each set of experiments. We find that optimizing over more checkpoints (8 and 16) outperforms optimization over fewer checkpoints (2,4), showing how recycling more models can outperform recycling fewer checkpoints.}
    \label{fig:n_ckpts_exp_mbpp_ifeval}
\end{figure}

\Cref{tab:triwise} shows results from our three-task experiment in \Cref{sec:three-task} and \Cref{fig:n_ckpts_exp_mbpp_ifeval} includes results with search over different number of checkpoints over MBPP-IFEval from \Cref{sec:analysis}.

\begin{table*}[ht]
\centering
\footnotesize
\begin{NiceTabular}{@{}lccccccc@{}}[colortbl-like]
\toprule
\textbf{Model} & \multicolumn{4}{c}{\textbf{Held-in}} & \multicolumn{2}{c}{\textbf{Held out}} & \textbf{Avg. All Tasks} \\ 
\cmidrule(lr){2-5} \cmidrule(lr){6-7} \cmidrule(lr){8-8}
& \textbf{MBPP} & \textbf{IFEval} & \textbf{GSM8K} & \textbf{Avg.} & \textbf{MT-Bench} & \textbf{LBPP} & \\ \midrule

\rowcolor{cyan!15} Highest fitness model  & 56.8 & 72.0 & 81.0 & 69.9 & 7.42 & 30.4 & 49.52 \\
\rowcolor{cyan!15} Best on MBPP &  64.0 & 56.5 & 75.7 & 65.4 & 7.68 & 32.9 & 47.36 \\ 
\midrule
\rowcolor{orange!20} Uniform Soup & 62.4 & 68.2 & 79.5 & 70.0 & 8.24 & 32.3 & 50.13 \\
\rowcolor{orange!20} Merge best & 62.2 & 69.3 & 80.5 & 70.7 & 8.16 & 32.3 & 50.49 \\ 
\rowcolor{orange!20} Optimized Merge & 63.6 & 71.9 & 80.9 & \textbf{72.1} & 8.21 & 33.5 & \textbf{51.62} \\

\bottomrule
\end{NiceTabular}
\caption{Comparison of model performance across different task pairs. Held-in tasks refer to tasks included in the fitness function (\S~\ref{sec:method-optimization}).}
\label{tab:triwise}
\end{table*}

\begin{figure*}[t!]
    \centering
    \includegraphics[width=0.32\linewidth]{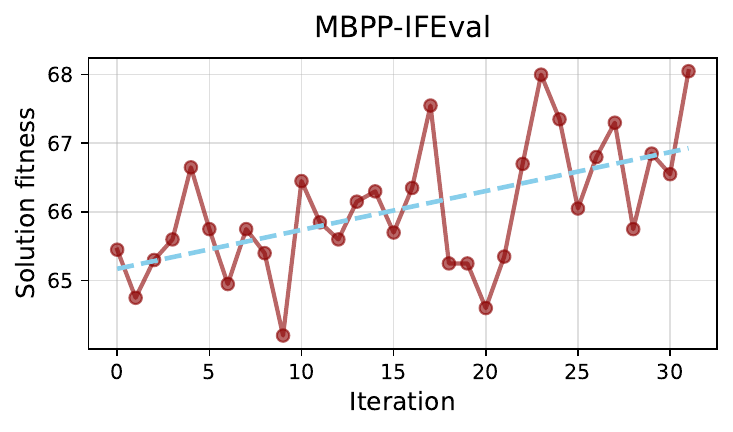}
        \includegraphics[width=0.32\linewidth]{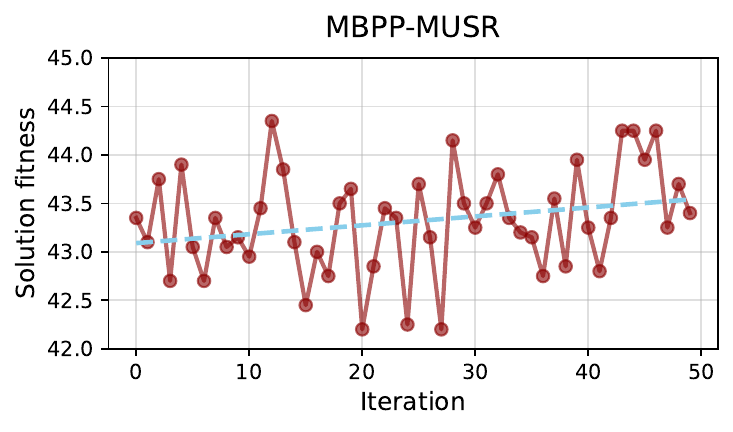}
        \hfill
    \includegraphics[width=0.32\linewidth]{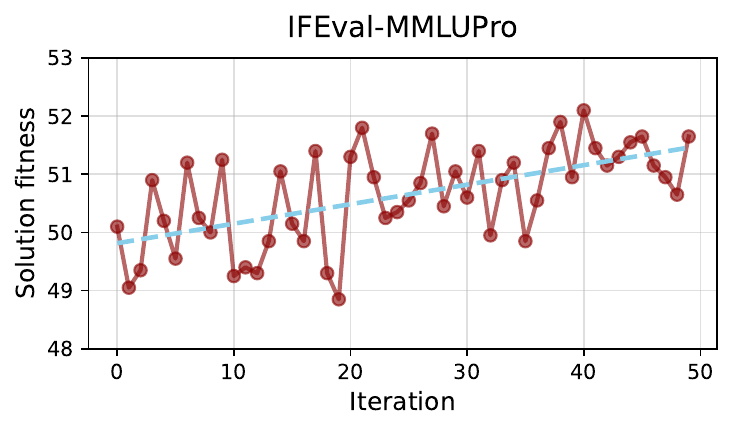}
    \caption{Fitness vs. CMA-ES iterations when optimizing tradeoffs over task pairs (see \Cref{sec:pairwise-tradeoffs}). CMA-ES explores the search space to find merge weightings with high fitness or low task tradeoffs.}
    \label{fig:search-prog}
\end{figure*}

\Cref{fig:search-prog} shows fitness improvement over the course of CMA-ES. While the improvement in the fitness is not monotonic due to the sampling nature of CMA-ES, the average fitness shows a positive trend with more iterations.

\section{Computational Cost Comparison}
\label{app:search-cost}
Following \citet{kaplan2020scaling}, the total training cost in FLOPs is estimated by:
\[
\text{Train FLOPs} \;=\; 6\,N\,B\,S,
\]
where \(N\) is the number of non-embedding parameters, \(B\) is the batch size, and \(S\) is the number of training steps. In our case, \(N \approx 10^{11}\), so we can estimate the cost of a single stage of supervised finetuning (SFT) and preference optimization (PO) training stages cost as follows:

\[
\begin{aligned}
&\text{SFT:} \quad 6 \times 100\times10^{9} \times 64 \times 1554 \;=\; 6\times10^{16}, \\
&\text{PO:} \quad 6 \times 100\times10^{9} \times 64 \times 1182 \;=\; 4.57\times10^{16}, \\
&\text{Total:} \quad 6\times10^{16} + 4.57\times10^{16} \;=\; 1.057\times10^{17}.
\end{aligned}
\]

In contrast, inference costs are substantially lower. From the same scaling laws paper, inference takes about
\[
\text{Inference FLOPs} \;=\; 2\,N \times \text{\#samples},
\]
leading to the following cost (with \(N \approx 10^{11}\)) on different tasks:

\[
\begin{aligned}
&\text{MBPP:} \quad 2 \times 100\times10^{9} \times 500 \;=\; 1.01\times10^{14},\\
&\text{IFEval:} \quad 2 \times 100\times10^{9} \times 541 \;=\; 1.09\times10^{14},\\
&\text{MTBench:} \quad 2 \times 100\times10^{9} \times 80 \;=\; 1.61\times10^{13},\\
&\text{GSM8K:} \quad 2 \times 100\times10^{9} \times 1300 \;=\; 2.6\times10^{14}, \\ 
&\text{MUSR:} \quad 2 \times 100\times10^{9} \times 756 \;=\; 1.512 \times10^{14},\\ 
&\text{LBPP:} \quad 2 \times 100\times10^{9} \times 161 \;=\; 3.24\times10^{13}, \\
&\text{MMLUPro (2K):} \quad 2 \times 100\times10^{9} \times 2000 \;=\; 4.0\times10^{14}
\end{aligned}
\]

Since we run the search for 50 iterations, the total compute cost for a full merge optimization over two tasks (e.g., MBPP and IFEval) amounts to:  

\[
50 \times 1.01 \times 10^{14} + 50 \times 1.09 \times 10^{14} = 1.05 \times 10^{16} \text{ FLOPs}.
\]

That means that optimized model merging only needs at most 10\% compute as that needed for a single stage of SFT + PO training. In practice, multiple SFT stages are applied over many training runs to explore different hyperparameter choices, further amplifying the overall training cost compared to search-optimized merging. 

\end{document}